\documentclass[10pt,twocolumn,letterpaper]{article}

\usepackage{cvpr}
\usepackage{times}
\usepackage{epsfig}
\usepackage{graphicx}
\usepackage{amsmath}
\usepackage{amssymb}
\usepackage{float}
\usepackage{subfigure}

\usepackage[pagebackref=true,breaklinks=true,letterpaper=true,colorlinks,bookmarks=false]{hyperref}

\cvprfinalcopy 

\def\cvprPaperID{****} 

\begin{document}

\title{Out of the Box: A combined approach for handling occlusions \\in Human Pose Estimation \\ {\small{Seminar Report}}}

\author{Rohit Jena\\
Computer Science and Engineering\\
Indian Institute of Technology, Bombay\\
{\tt\small rohitrango@cse.iitb.ac.in}
}

\maketitle

\begin{abstract}
Human Pose estimation is a challenging problem, especially in the case of 3D pose estimation from 2D images due to many different factors like occlusion, depth ambiguities, intertwining of people, and in general crowds. 2D multi-person human pose estimation in the wild also suffers from the same problems - occlusion, ambiguities, and disentanglement of people's body parts. Being a fundamental problem with loads of applications, including but not limited to surveillance, economical motion capture for video games and movies, and physiotherapy, this is a interesting problem to be solved both from a practical perspective, and from an intellectual perspective as well. Although there are cases where no pose estimation can ever predict with 100\% accuracy (cases where even humans would fail), there are several algorithms that have brought new state-of-the-art performance in human pose estimation in the wild. We look at a few algorithms with different approaches, and also formulate our own approach to tackle a consistently bugging problem, i.e. occlusions.  
\end{abstract}

\section{Introduction}
Human Pose Estimation is a fundamental, yet hard problem which aims to find one or more humans in real-time or non-real-time settings (in 2D or 3D). There are many applications related to vision-based human pose estimation (in contrast to sensor based approaches). Some of these are:
\begin{itemize}
  \setlength\itemsep{0pt}
    \item Markerless Motion Capture for human-computer interactions
    \item Physiotherapy
    \item Visual surveillance
\end{itemize}
In this report, I'm going to explore some of the current methods used in Deep Learning for the problem of human pose estimation. We begin by looking into the categories of human pose estimation, look into the overview of the algorithms of a few of them, and then formulate a very simple algorithm to take care of occlusions.

\section{Single Person 2D Pose Estimation}
Single Person pose estimation, as the name suggests, detects only a single person in an image. The problem is relatively easier in the sense that there is only one person to take care of, and if the image is constrained to contain a single person, then the problem essentially reduces to finding the joints of the person and then connecting them is trivial since we are given a skeleton structure of the person. It gets considerably harder to account for multiple people, because of false positives in case of joints being too close, and association of the keypoints. \\
However, detecting keypoints is not as easy of a task as it seems. With classical hand-engineered features like Histogram of Gradients (HOG), Difference of Gaussians, cluster-based methods etc., there are obvious limitations like unseen examples, lack of capturing global context, occluded joints (which will result in different detections and hence a false negative). But with the advent of Deep Learning, detections and localization problems have been incredibly simple given we have enough data to learn. We explore the first architecture for single-person detection, which is a Stacked Hourglass Network. 

\subsection{Stacked Hourglass Network}
The Stacked Hourglass network is a powerful model for single-person detection. The model uses Convolutional Neural Networks (CNNs) to learn abstract features on a local scale. The CNN has a receptive field - which is usually small and is the same as the size of the filter. Stacking these convolutions increases the receptive field, which increases the effective receptive field, and hence, captures more global context. Since convolutional networks don't need a fixed input size, the Hourglass utilizes a Fully Convolutional Network (FCN) to account for different image sizes and aspect ratios. FCNs are used for object detection, semantic segmentation, etc. 

\subsubsection{Architecture}
An hourglass is a stack of residual units which are stacked in an encoder-decoder fashion. The architecture is set up as follows: Convolutional and Max-pooling layers are used to process features and bring them to a lower resolution. Since global context is captured, the low-res images contain big patch level information. It also uses skip connections to use lower-level features while upsampling and helps in better flow of gradients.
\begin{figure}[t]
    \centering
    \includegraphics[width=0.9\linewidth]{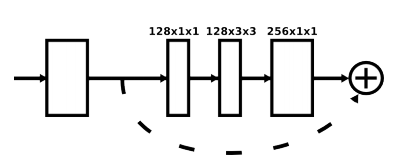}
    \caption{A residual block (Figure: \cite{stacked})}
    \label{fig:resblock}
\end{figure}

\begin{figure}[t]
    \centering
    \includegraphics[width=\linewidth]{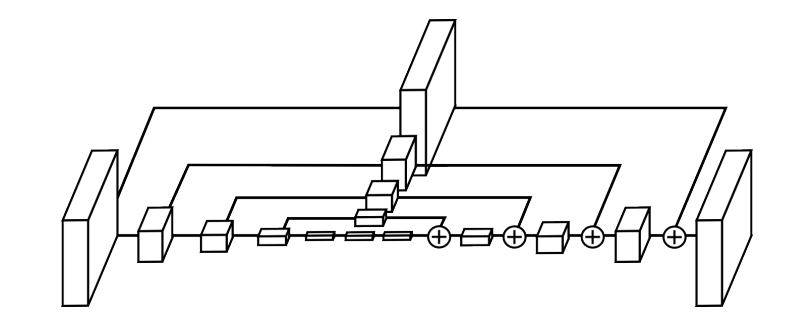}
    \caption{The hourglass model (Figure: \cite{stacked})}
    \label{fig:hourglass}
\end{figure}
The architecture (Figure:\ref{fig:hourglass}) is heavily inspired from an encoder-decoder network. They also employ other tricks to improve performance, for example, instead of using a 5x5 convolution, they replace it with two 3x3 convolutions. This also reduces the space and increases feature complexity. Recent work has shown the value of reduction steps with 1x1 convolutions, as well as the benefits of using consecutive smaller filters to capture a larger spatial context. \\
However, all the discussion is with respect to a single hourglass. Where does stacked hourglass come into the picture? And does it only make the network learn ``deeper'' features? The answer is no, and the stacking actually enables us to do intermediate supervision.
\subsubsection{Intermediate supervision}
Since the output of a hourglass is the same as the image, we can use the final layer of the hourglass to predict the keypoints, and convert them into features before passing it as input into the next hourglass. This makes the model learn the outputs early on, and the model tries to refine its predictions from layer to layer.
\begin{figure}[t]
    \centering
    \includegraphics[width=\linewidth]{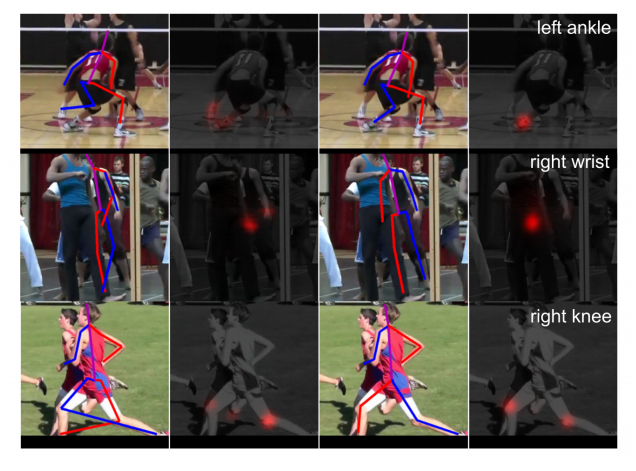}
    \caption{Refining the predictions using intermediate supervision (Figure: \cite{stacked})}
    \label{fig:predictrefine}
\end{figure}
Note that they do not "cheat" in the sense that if they have more hourglasses, then the depth of the hourglass is reduced, so as to have the same number of parameters. It is observed that more intermediate supervision helps in better refinement. Here is a diagram that describes the overall architecture:
\begin{figure}[t]
    \centering
    \includegraphics[width=\linewidth]{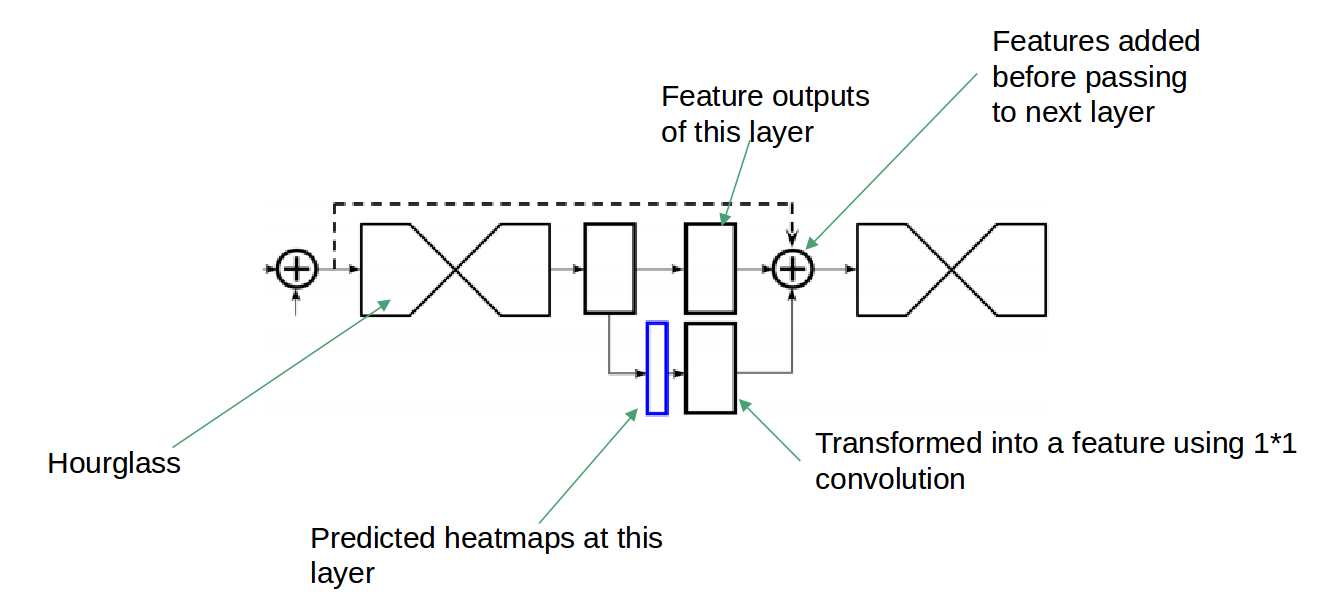}
    \caption{An overview of the intermediate supervision}
    \label{fig:intermsup}
\end{figure}
The limitation of this model arises when the image contains more than 1 person. Slight changes to the alignment of the image leads to a completely different pose estimation. In other words, there is no explicit ``focus'' on a single person with multiple people around. Hence, it is only desirable in situations where we want to explicit track only a single person, like in sensorless mo-cap systems in 2D animation. \\
All these features give a very powerful model for learning human pose keypoints and then associating the points is trivial since there is only 1 point for each joint. Also, this is easier than its 3D counterpart since 3D pose estimation also means accounting and estimating depth, relative ordinal depth relations, etc. It is also easier than 2D multi-person pose estimation because of overlap, entangled/intertwined people, and ambiguity in general. We look at 2D multi-person approaches and analyze their limitations.

\section{Multi-Person 2D Pose Estimation}
2D multiperson pose estimation adds a layer of possibilities and challenges along with it. 2D pose estimation may be desirable to track people ``in the wild'', maintain surveillance, security, etc. The problem extends to finding multiple people, and associating their joints to detect multiple poses. An ideal multi-person pose estimation algorithm should be person-agnostic in time complexity, i.e. it should take the same time to predict all the poses regardless of the number of the people in the image. However, current algorithms are not person-agnostic. Multi-person 2D Pose estimation algorithms are also divided into 2 categories depending on the approach:
\begin{itemize}
    \item \textbf{Top Down}: A top-down approach uses an initial detection step (bounding box, segmentation) and uses the result of this step to run a single person detector on top of it. Popular top-down approaches are Mask RCNN, LCRNet, etc
    \item \textbf{Bottom Up}: A bottom-up approach uses an initial detection step to find out \textbf{all} joints, and some other association parameters and structural information. The second step uses a matching algorithm to utilize the structural information and the potentially multiple keypoints for a given joint to detect all poses.
\end{itemize}
Both the algorithms have an $O(n)$ step, where $n$ is the number of people. For the top-down, it is running the single person detector for $n$ people. For bottom-up, it is proportional to the number of joints detected, or possibly even a polynomial time in $n$. Both the approaches have their own strengths and weakness, and we discuss them after discussing their corresponding algorithms. \\

\subsection{Mask-RCNN}
Mask-RCNN \cite{maskrcnn} is of the most popular algorithms for object detection, segmentation, and keypoint detection. 
\subsubsection{Algorithm}
The algorithm is conceptually very simple: Given an image, a base network first predicts ``object proposals'' which are boxes within the image where the network thinks an object may be. This doesn't tell the class of the object yet, and the boxes are predicted just on the ``objectness'' in the box. The second part is to use a cropped feature map from the image (according to the box) which predicts class of the object and bounding box offsets for refinement of boxes. A third branch also predicts binary pixelwise masks for instance segmentation. 

\subsubsection{Architecture}
This is the general framework of the Mask RCNN architecture. The head of the network is typically a ResNet-50 architecture. The aim of the network is to output general task-invariant features which can be shared across the classification, regression, and segmentation branches. 
\begin{figure}[t]
    \centering
    \includegraphics[width=\linewidth]{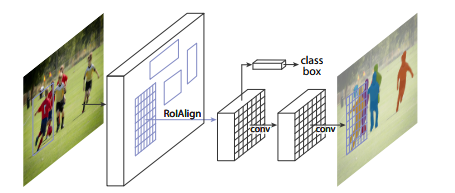}
    \caption{Mask-RCNN architecture (Figure: \cite{maskrcnn})}
    \label{fig:mrcnn}
\end{figure}
The bounding boxes are then cropped by an RoIAlign step, which crops images from the bounding box using bilinear interpolation for sampling to avoid information loss due to nearest neighbour sampling, which helps in more accurate segmentation. 
\begin{figure}[t]
    \centering
    \includegraphics[width=\linewidth]{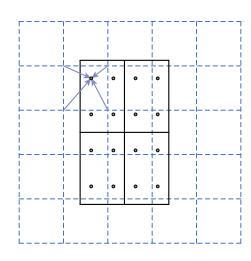}
    \caption{The RoIAlign step. Notice how the values are filled in using the 4 nearest points from the feature map, and the weights being the ones used in bilinear interpolation, instead of nearest neighbour sampling, (which is essentially RoIPool) which introduces artifacts in both the sampled image and the resulting segmentation \cite{maskrcnn}}
    \label{fig:roialign}
\end{figure}
For keypoints detection, we introduce yet another branch apart from classification, regression and segmentation branches, which is the keypoint branch. For K joints, the branch predicts K-mxm masks, each for one keypoint. During training, for each visible ground-truth keypoint, the cross-entropy loss is minimized over an $m^2$-way softmax output. This encourages the network to predict a single keypoint for each mask. In the base network, the FPN variant is commonly used to detect at multiple scales more effectively. This is an example of a top-down inference architecture. Now, we look at a different approach, called Part Affinity Fields. \\

\subsection{Part Affinity Fields}
Part Affinity Fields (PAFs) are a complementary approach to the Detectron framework. In Detectron, we go the top-down way, which is to first detect bounding boxes and then detect single person keypoints. However, sometimes bounding boxes may capture less than one person (missed joints) or sometimes it may capture the person but also other people who are very close. Part Affinity Fields are a non-parametric bottom-up approach, which detects all the keypoints (heatmaps essentially) for all the people simultaneously. However, the magic of the algorithm lies in the association step.

\begin{figure}[t]
    \centering
    \includegraphics[width=0.9\linewidth]{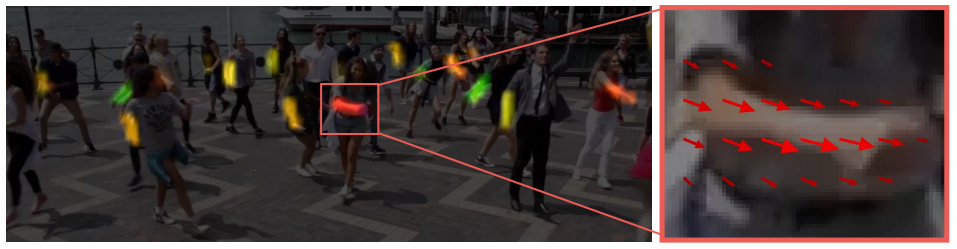}
    \caption{Part Affinity Fields (Figure: \cite{paf})}
    \label{fig:paf}
\end{figure}

The structural information of the people's joints and their relationships are predicted by 2D vectors at each pixel, which defines the ``affinity of the parts being part of a limb'' and aptly named Part Affinity fields, field because the directed vectors are output at a per-pixel level, and hence, is analogous to a force or electric field. This approach tackles the Early Commitment Problem, which is basically the problem that once a bounding box is decided, the detector will predict only inside the box, and hence, depends a lot on the accuracy of bounding boxes. It tackles it because every step of the inference is done at the whole image level, so nothing is missed out unless in case of a false-positive or false-negative keypoint/fields. With lots of training data, predicting keypoints with high accuracy is possible for the whole image. We now discuss it's architecture and working.

\subsubsection{Architecture}
The architecture is divided into 2 branches - the Keypoints branch and the Part Affinity Branch. Given an image feature \textbf{F} of size $hxw$, which is usually the output of a ResNet architecture, we predict joint confidences \textbf{S} and PAFs \textbf{L}. To enable multi-stage learning, the confidences and part affinity fields are concatenated with the features as input to the next stage. This helps in intermediate supervision as well.

\begin{figure}[t]
    \centering
    \includegraphics[width=0.95\linewidth]{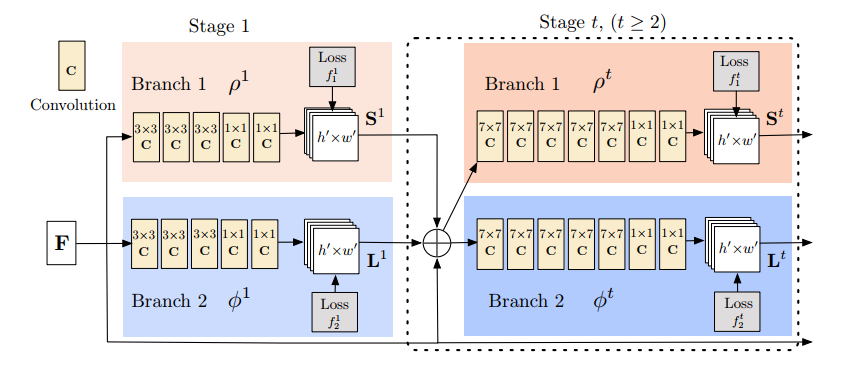}
    \caption{Part Affinity Fields - Architecture (Figure: \cite{paf})}
    \label{fig:archpaf}
\end{figure}

The set $\bf{S = \{ S_1, S_2, .... S_K \}}$ has $K$ confidence maps, each for one of the $K$ joints. Note that for a given joint map, there can be multiple predictions, each for a different person. Hence, we use a sigmoid activation instead of a softmax, which encourages just a single maximum probability of getting a joint estimation. We have $S_j \in \mathbb{R}^{wxh} \quad \forall j$.
The set $\bf{L = \{ L_1, L_2, .... L_C \}}$ has $C$ vector fields, one for each limb. A limb is defined as a pair of joints that are directly connected via an edge, given a defined skeleton of the human pose. Since we want 2D vectors at every pixel, $L_c \in \mathbb{R}^{wxhx2}$. For training, the ground truth is constructed in the following way: For the confidence maps $S^*$, we first construct person-wise confidence maps $S_{j,k}^*$ for joint j, and person k. The value at a pixel \textbf{p} is given by
$$ S_{j,k}^*(p) = \exp\Bigg(-\frac{||p-x_{j,k}||_2^2}{\sigma^2}\Bigg)$$
where $\sigma$ controls the spread of the peak. The final confidence map for combining all the people's confidence map is given by 
$$ S_j^*(p) = \max_kS^*_{j,k}(p)$$
The max is used instead of average because we want to preserve the confidences of different people who are close to each other. Averaging them out will modify their peaks, and lead to inaccurate training. The Part Affinity Fields are evaluated for every pair of joints. To define the groundtruth part affinity fields we use the person-wise PAF as follows:
$$ L^*_{c, k}(p) = 
\begin{cases}
v & {\text{if p on limb \textit{c,k}}} \\
0 & {\text{otherwise}}
\end{cases}
$$
Here, $v$ is defined as $\frac{x_{j_2,k} - x_{j_1, k}}{||x_{j_2,k} - x_{j_1, k}||}$, which is the unit vector in the direction of the limb. The set of points on the limb is defined as those within a distance threshold of the line segment, i.e., those points $p$ for which
$$ 0 \leq v.(p - x_{j_1, k}) \leq l_{c, k} \quad\text{and}\quad |v_{\perp}.(p - x_{j_1, k})| \leq \sigma_l $$
The ground truth Part Affinity Field is obtained by averaging the part affinity fields of all the people 
$$ L_c^*(p) = \frac{1}{n_c(p)}\sum_k L_{c, k}^*(p)$$
where $n_c(p)$ = count of all the people whose non-zero vectors were considered. If $n_c(p) = 0$, then $L_c^*(p) = 0$. 
During training, the loss function is the following (note that we had multi-stage training, so the loss functions are given for each stage, say stage \textit{t})
$$ f^t_S = \sum_{j=1}^{J}\sum_p W(p).||S^*_j(p) - S^t_j(p) ||^2_2$$
$$ f^t_L = \sum_{j=1}^{J}\sum_p W(p).||L^*_j(p) - L^t_j(p) ||^2_2$$
Where $W$ is a binary mask, which is 0 where there is no annotation for any people. The importance of $W$ is that there may be people whose annotations may not be included in the ground truth annotations. We do not want the network to penalize if these poses are also detected. Hence, the mask $W$ makes sure we only flow gradients through the annotations we know of. 
\subsubsection{Inference}
For inference, we first detect the keypoints using the heatmaps. We find all points which are a local maxima, and suppress repeated peaks using NMS. After finding the points, we take all possible point pairs for a given limb, say $\bf{d_{j_1}}$ and $\bf{d_{j_2}}$. We calculate the score of association between these points as follows: 
$$ E = \int_{u=0}^{u=1}{\bf{L}}_c(p(u)){\bf{\frac{d_{j_1} - d_{j_2}}{||d_{j_1} - d_{j_2}||}}} \mathrm{d}u$$
where $p(u) = (1-u){\bf{d_{j_1}}} + u{\bf{d_{j_2}}}$. We approximate the integral by sampling points from the line segment connecting the 2 points. In the paper, they take 10 equally spaced points. Once they have a score, they use \textit{bipartite matching} algorithm for every pair of limbs to find the maximum scored limbs. Then the limbs are connected to output final poses. Note that this is a greedy strategy since we use locally optimum solutions (i.e. limbs) to find global optimum pose. However, this algorithm works good in practice. 

\subsubsection{Limitations}
Top-down approaches are better in the sense that given a bounding box, the keypoint detector is very accurate since it captures contextual and structural information inherently, as compared to bottom-up approaches where the structural information is indirect and a matching algorithm can be NP-hard in time complexity. However, the bounding boxes detect the ``objectness'' of the box, so any extension to the box which brings other objects restricts the bounding box, and subsequently the pose estimation. Sometimes extending the bounding box to accommodate a part may not be enough, because that results in parts of other people, which leads to confusion for the single person detector. Although bottom-up approaches are resistant to occlusion, they can mess up in case of complex poses or intertwined people (especially if the association step is greedy), since the per-pixel structural information is now averaged out across people. Also, these approaches are not very accurate for small people. \\
Now, we move to another section, which is the problem of inferring 3D poses from 2D images, also called 3D human pose estimation.

\section{Single Person 3D Pose Estimation}
Estimating 3D pose from 2D images is an ambiguous problem in itself due to the complexities of inferring depth of visible and hidden joints, inferring positions of hidden joints by having a prior on human joints, or using other methods. There is one more problem which 3D pose estimations suffer from - not having enough 3D annotations to learn from. Most 3d human datasets are in controlled environments (like the Human 3.6m dataset), and hence ``in the wild'' 3D pose estimation requires other techniques like single-shot learning, unsupervised learning methods, etc. We look at some single person 3D pose estimation algorithms at first, and then move on to another popular multiperson 3D pose estimation method. These are most applicable in person tracking for markerless applications. We discuss two frameworks - VNect, and Geometry-Aware Pose estimation.

\subsection{VNect}
VNect \cite{vnect} is an approach to single-person 3D human pose estimation. The idea is to use 2D RGB monocular images, and use previous frame information for refining the poses and handling temporal jitter as well. The approach is very similar to the Mask RCNN approach described before. However, there are a few differences in terms of how the 3D pose is calculated. The pipeline consists of a bounding box step, followed by heatmap predictions, and in addition, they have location maps. A temporal filter takes in the keypoints and pose information from the previous frame, and helps in refining the pose of the current frame. This is followed by a temporal skeleton fitting in 3D.

\subsubsection{Architecture}
\begin{figure*}[t]
    \centering
    \includegraphics[width=\linewidth]{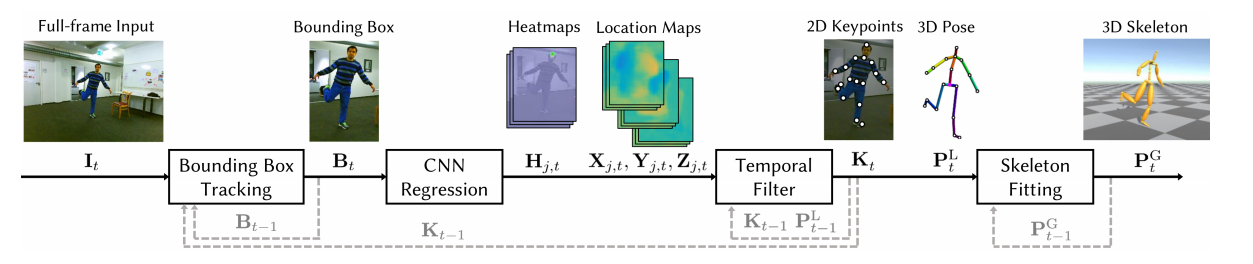}
    \caption{VNect - Architecture (Figure: \cite{vnect})}
    \label{fig:vnect}
\end{figure*}
The overview of the architecture of the VNect is given in \ref{fig:vnect}. The main parts of the architecture are the following:
\begin{itemize}
    \item \textbf{Bounding box tracking}: This is an initial step that makes a bounding box around the person to crop out unnecessary clutter. This is similar to mask RCNN except that we have a single bounding box prediction, and a bounding box tracker also uses the box from the previous frame to keep consistency between the immediate predictions. The 2D pose predictions from the CNN at each frame are used to determine the bounding box for the next frame through a slightly larger box around the predictions. The smallest rectangle containing the keypoints $K$ is computed and augmented with a buffer area which 0.2x the height vertically and 0.4x the width horizontally. To stabilize the estimates, the bounding box is shifted horizontally to the centroid of the 2D predictions, and its corners are filtered with a weighted average with the previous frame’s bounding box using a momentum of 0.75. 
    \item \textbf{CNN regression}: The next step is to use a FCN based architecture on this image to obtain heatmaps of the keypoints individually. Instead of direct regression, heatmaps provide a more intuitive mapping from the image, and are easier to learn. These heatmaps give the 2D position of the person, which we call $K_t$. 
    \item \textbf{Location maps}: To get the 3D pose $P_t^L$ from the 2D keypoints, we output something called location maps. These location maps, at each pixel, give the x, y, and z coordinates of the 3D pose $P_t^L$ relative to the root. The 3D coordinates $(x, y, z)$ of a joint $j$ is obtained from the coordinates corresponding to the 2D joint location. 
    \item \textbf{Temporal Filtering}: The previous $K_{t-1}$ and $P^L_{t-1}$ are used to refine the keypoint and pose for the current time frame. 
\end{itemize}
For learning the heatmap, we follow the approach discussed in previous approaches, i.e. Mask-RCNN, Part Affinity Fields. To learn the location maps $X_{j,t}, Y_{j,t}, Z_{j,t}$, we use a similar loss function that we used to learn the Part Affinity fields. To show that we are only interested in the locations $x_j, y_j, z_j$, we use the following loss function:
$$ L(x_j) = ||H^{GT}_j\odot(X_j - X^{GT}_j) ||_2$$
Where, L2 loss is used for norm. We note that the matrix $H^{GT}_j$ is similar to the $W$ matrix, which only allowed gradients through the points we are interested in. However, the purpose is quite different. 

\begin{figure*}[t]
    \centering
    \includegraphics[width=\linewidth]{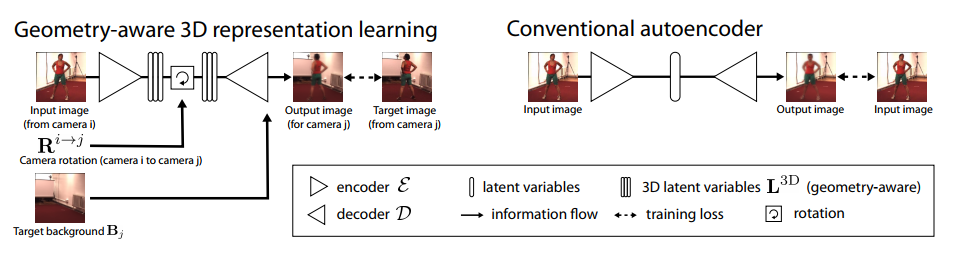}
    \caption{Representation Learning (Figure: \cite{aware})}
    \label{fig:reprlr}
\end{figure*}

\subsubsection{Kinematic Skeleton Fitting}
Applying independent pose estimation for 2D and 3D may introduce temporal jitter since there is no information from previous frames, and deep networks may not be familiar with a frame which is a bit out of the manifold. Hence, a kinematic skeleton fitting step is used to factor these out. The 2D predictions $K_t$ are temporally filtered and used along with the location maps to obtain 3D pose $P^L_t$. The bone lengths inherent to $P^L_t$ are replaced by the bone lengths of the skeleton in a retargeting step that preserves the bone directions of $P^L_t$. The 2D and 3D poses are combined using the following energy:
$$ E_{\text{total}}({\bf \theta, d}) = E_{\text{IK}}({\bf \theta, d}) + E_{\text{proj}}({\bf \theta, d})$$ 
$$ + E_{\text{smooth}}({\bf \theta, d}) + E_{\text{depth}}({\bf \theta, d})$$
where ${\bf\theta}$ are the joint angles, which contain information about the pose (changing joint angles will obviously change the pose) and ${\bf d}$ are the root joint's location on camera. The individual energy terms are given by:
$$E_{\text{proj}}({\bf \theta, d}) = ||\small\prod P^G_t - K_t||_2$$
This term penalizes the projection of the 3D pose into the 2D image space, and the keypoint detections in the 2D image. So the 3D pose cannot result in anything which leads to a different 2D pose.
$$E_{\text{IK}}({\bf \theta, d}) = ||(P^G_t - d) - P^L_t ||_2$$
This term makes sure that the 3D pose is not too different from the pose predicted by the location maps and 2D keypoints. 
$$E_{\text{smooth}}({\bf \theta, d}) = ||\hat{P^G_t}||_2$$
The smoothing term penalizes acceleration in the 3D keypoints.
$$E_{\text{depth}}({\bf \theta, d}) = ||{[\tilde{P}^G_t]_z}||_2$$
This term large variations in depth temporally, where ${[\tilde{P}^G_t]_z}$ is the z-component of velocity of the 3D pose.
Minimization of these terms gives a temporally consistent 3D pose. Now, we discuss another algorithm which uses unsupervised learning to learn 3D poses from a latent 3D pose space, and requires very little 3D annotated data.
\subsection{Geometry-Aware Pose representation}
The main concern of the paper is that we are now limited by quality of datasets. Since annotation of 3D pose is both ambiguous and expensive, methods that use weak supervision are favorable. To make use of a lot of unlabeled 3D data, the training of human pose estimation is done in 2 steps. In the first step, a geometry-aware representation is learnt. This geometry aware representation delineates appearance, background, and the 3D pose in a latent space. This is done only using unlabeled 3D data. In the second step, the latent 3D pose representation is mapped to an actual 3D pose. This requires limited data, and hence a small network is required to be learnt for this step, in contrast to deep regressors.
\subsubsection{Latent Pose Representation}
The goal is to learn a latent representation and use it for 3D pose estimation. For the representation to be practical, they break the latent representation into 3 parts: ${\bf L_{app}, L_{3D}, B}$, corresponding to the appearance, 3D pose, and background respectively. The architecture (\ref{fig:reprlr}) is heavily inspired from the encoder-decoder architecture, which learn a latent representation of the input. However, the learning here is modified so that the appearance, pose, and background are all delineated from each other. Let $\mathcal{U} = {({\bf I}_t^i, {\bf I}_t^j)}_{t=1}^N$ be pairs of images from different camera views $i$ and $j$. Let $R_{i\rightarrow j}$ be the rotation matrix for rotation from camera $i$ to camera $j$. The 3 terms are learned in the following way:
\begin{itemize}
    \item \textbf{Factoring background}: The background for a view $j$ is calculated by computing the median of all frames from camera $j$. This makes sense, because the movements of a person would be factored out due to a large number of frames containing some portion of the background.
    \item \textbf{Factoring pose}: This is the main step of interest. To factor out pose, we consider two views $i$ and $j$ of the same image at the same time $t$. To learn a semantic pose representation, we first observe that the rotation matrix $R_{i\rightarrow j}$ is a 3x3 matrix. So, we model the $L_{3D}$ as a 3xN matrix, so that $R_{i\rightarrow j}.L_{3D}$ is also a 3xN matrix, and it follows the rotation semantics. So, we input the image $I_i^t$ to the encoder, $\mathcal{E}(I_i^t) = L_{3D, i}^t, L_{app, i}^t, B_i^t$, and we use the $R_{i\rightarrow j}.L_{3D, i}$ as input to the decoder, giving 
    $\hat{I} = \mathcal{D}(R_{i\rightarrow j}.L_{3D, i}, L_{app, i}^t, B_i^t)$ and then minimizing $||I_j^t - \hat{I}_j^t||_2$. This enforces $R_{i\rightarrow j}.L_{3D, i}$ to be close to $L_{3D, j}$.
    \item \textbf{Factoring appearance}: To factor out the appearance, we take 2 time frames $t$ and $t'$ with $t \neq t'$ and $|t - t'| < K$. An assumption here is that the person has the same appearance, and hence while learning, the encoder outputs $\mathcal{E}(I_j^t) = L_{3D, j}^t, L_{app, j}^t, B_j^t$. But for the decoder we input $L_{3D, j}^t, L_{app, j}^{t'}, B_j^t$ and minimize $||I_j^t - \hat{I}_j^t||$ where $\hat{I}_j^t = \mathcal{D}(L_{3D, j}^t, L_{app, j}^{t'}, B_j^t)$. This makes the encoder output only the appearance, as we have already delineated the pose information before.
\end{itemize}
Hence, we combine all of these parameters jointly, and train the autoencoder with the following :
$$ \mathcal{A}(I_t^i, R_{i\rightarrow j}, K^{app}_{k, t'}, B_j) = \mathcal{D}(R_{i\rightarrow j}L^{3D}_{i, t}, L^{app}_{k, t'}, B_j) $$
The combined optimization is as following: to optimize over rotation, appearance, and background simultaneously, mini-batches of $I_{(t, i)}$, $I_{(t, j)}, I_{(t’, i)}$ where $i\neq j$ and $t \neq t'$ are picked and the following objective function is trained:
$$ E = \frac{1}{Z}\sum_{I^i_t, I^j_t, I^k_{t'} \in \mathcal{U}} ||\mathcal{A}(I_t^i, R_{i\rightarrow j}, K^{app}_{k, t'}, B_j) - I^j_t||_2 $$
\begin{figure*}[t]
    \centering
    \includegraphics[width=\linewidth]{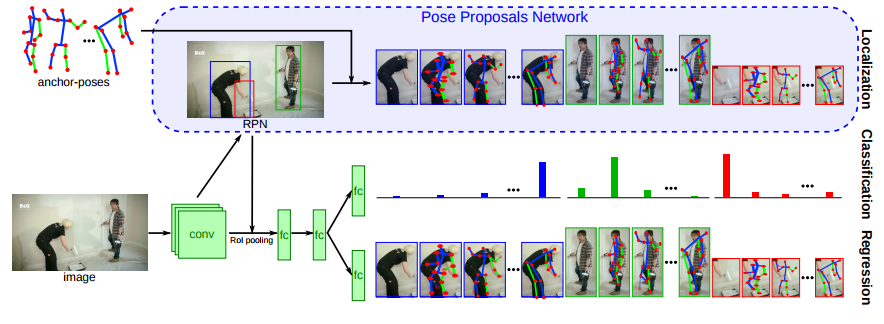}
    \caption{Architecture of LCRNet++ (Figure: \cite{lcr})}
    \label{fig:lcrnet}
\end{figure*}
\subsubsection{Learning 3D poses}
Now that we have trained the autoencoder-type architecture using lots of unsupervised data, the supervised part is when we have a few annotations of image and pose, we take the latent 3D pose as input, and predict the keypoints as output. This requires a very small network to learn, and this is done in the 2nd stage after learning the latent pose. Hence, given an input image $I^i_t$, the following objective is trained:
$$ F_{\theta} = \frac{1}{N_s}\sum_{t=1}^{N_s}||\mathcal{F}_{\theta}(L_{3D}^t) - p||$$
where $L_{3D}^t$ is the 3D latent pose. Now, we discuss a popular multiperson 3D pose estimation framework, called LCRNet++. \\

\section{Multi-Person 3D Pose Estimation}
This is the final type of human pose estimation that is possible. This is also the most challenging type because it adds the complexity of 2D multiperson and 3D single person pose estimation together. A recent and popular network is LCRNet++, which aims at multi-person 2D and 3D human pose estimation. 
\subsection{LCRNet++}
LCRNet++ \cite{lcr} is an architecture that is aimed at multiperson 3D human pose estimation in a very similar way as that of RCNN. The architecture is shown in Figure \ref{fig:lcrnet}. Like the RCNN, the LCRNet++ network also consists of 3 parts, the first part being a Pose Proposal Network, which is basically an RPN with outputs bounding boxes for humans. Each bonding box is supposed to be associated with $K$ anchor poses, which will be defined later. These bouuding boxes are then cropped and passed into a classification and a regression branch. Here, the classes correspond to one of the $K$ anchor poses, and a 0 class for background. The regression task consists of $5J$ outputs for each of the $K$ anchor poses, the first two outputs being the deviation of the 2D coordinate (x,y) from the anchor pose, and the latter three outputs are the deviations from the 3D pose coordinates (x, y, z). Now we delve a little deeper into each of the branches. 
\subsubsection{Localization}
The localization branch is also called the Pose Proposal Network. The RPN predicts bounding boxes for the image, and the loss is given by
$$ \mathcal{L}_{loc} = \mathcal{L}_{rpn}$$
This is just an ordinary RPN, but at a conceptual level, we apply each of the anchor poses on the bounding box to get a total of NxK pose proposals for N bounding boxes.
\subsubsection{Classification}
The classification branch takes a cropped RoI as input, and predicts a probability distribution over $K+1$ classes, K classes for poses, and 1 class for background, which the RPN might have falsely added. The ground truth pose proposal is set for a bounding box, if it has an IoU > 0.5 with any ground-truth pose. If it has multiple intersections, the one with highest IoU is taken. Then, the anchor pose which has the least distance in 3D pose with the ground truth pose is taken as the ground-truth class for the box. The loss is given by:
$$ \mathcal{L}_{class}(u, c_B) = -\log(u(c_B)) $$
which is basically the normal negative log-likelihood loss. 

\begin{figure*}
    \centering
    \subfigure[]{\includegraphics[width=0.23\linewidth]{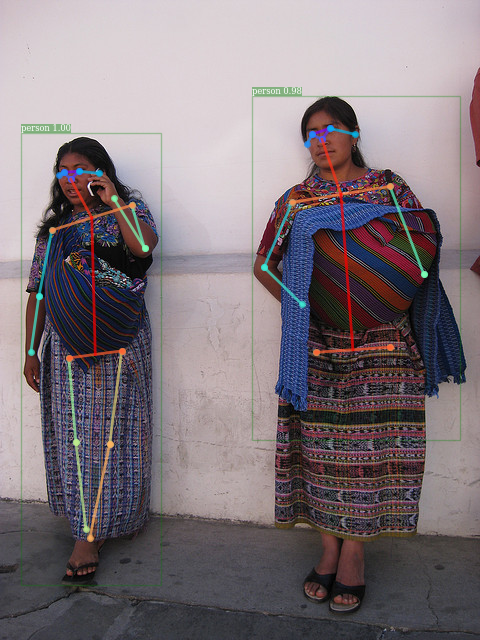}}
    \subfigure[]{\includegraphics[width=0.23\linewidth]{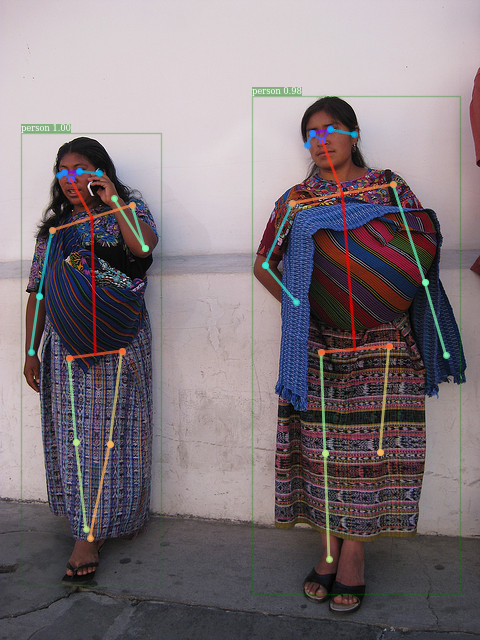}}
    \subfigure[]{\includegraphics[width=0.23\linewidth]{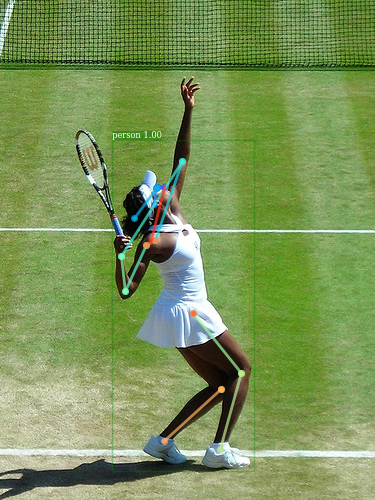}}
    \subfigure[]{\includegraphics[width=0.23\linewidth]{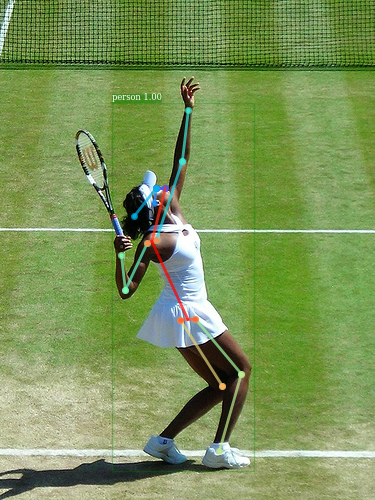}} 
    \caption{Note that our algorithm is suitably extending the bounding boxes, and adding more context even helps the pose estimator to refine the poses (images (c) and (d))}
\end{figure*}
\begin{figure*}
    \centering
    \subfigure[]{\includegraphics[width=0.48\linewidth]{wo_000000022935.png}}
    \subfigure[]{\includegraphics[width=0.48\linewidth]{000000022935.png}}   
\end{figure*}
\subsubsection{Regression}
The regression branch contains $5J(K+1)$ regression units, and for a ground truth class $c_B$, the ground truth regression task is given by $t_{c_B} = (\Hat{p} - \Hat{a_{c_B}}, \Hat{P} - \Hat{A_{c_B}})$. The regression loss is given by: 
$$ L_{reg}(v, t_{c_B}) = ||t_{c_B} - v_{c_B}||_s$$
where the norm is the Huber loss with $\lambda = 1$. The loss is applied only the subvector corresponding to the true class $c_B$. This is a general overview of the LCRNet++ architecture and training, as well as inference, since inference can be easily done by predicting class, and then the regression targets help in refining the pose. Now, we come to our approach.

\section{Going out of the Box}

\begin{table*}[t]
    \begin{minipage}{0.6\linewidth}
    \centering
    \begin{tabular}{|c|c|c|c|c|}
        \hline
        Metric & \textbf{Detectron} & \textbf{PAF test-dev} & \textbf{PAF test-challenge} & \textbf{Ours} \\ \hline
        AP & 0.6423 & 0.618 & 0.605 & \textbf{0.6465} \\ \hline
        AP50 &  0.8643 & 0.849 & 0.834 & \textbf{0.8677} \\ \hline
        AP75 &  0.6992 & 0.675 & 0.664 & \textbf{0.7044} \\ \hline
        APm &  0.5854 & 0.571 & 0.551 & \textbf{0.5904} \\ \hline
        APl &  0.7339 & 0.682 & 0.681 & \textbf{0.7361} \\ \hline
    \end{tabular}
    \vspace*{5pt}    
    \caption{Results on the COCO keypoints 2017 validation and test datasets}
    \label{tab:results}
    
    \end{minipage}
    \begin{minipage}{0.35\linewidth}
    \centering
    \begin{tabular}{|c|c|c|}
        \hline
        Metric & \textbf{Detectron} & \textbf{Ours} \\ \hline
        AP &  0.4759 & \textbf{0.477} \\ \hline
        AP50 &  0.7687 & \textbf{0.7696} \\ \hline
        AP75 &  0.539 & \textbf{0.541} \\ \hline
        APm &  0.4173 & \textbf{0.42} \\ \hline
        APl &  0.4973 & \textbf{0.4989} \\ \hline
    \end{tabular}
    \vspace*{5pt}
    \caption{Results on the MPII validation dataset} 
    \label{tab:resultsmpi}    
    \end{minipage}
\end{table*}

The name is pretty-much inspired from the way the Mask RCNN works by putting boxes and restricting the pose to be in the box. But in case of occlusions, this method fails. To alleviate this problem while keeping the structural information the keypoints detector intact, we introduce a new, simple, post-processing step which tackles some cases of occlusion, and in some cases, even refines the pose better due to more context available. The algorithm is as follows:
\begin{itemize}
    \item Predict the bounding boxes from the RPN step. 
    \item Given a defined directed tree of the skeletal structure of the human pose, we start from the root of the tree in a breadth-first order manner.
    \item For every joint $j$ in the PAF, try to expand the box it is in along a direction if there are no ``extra overlap'' with other boxes, and the direction fields point outward for the joint $j$.
    \item Feed the boxes into the person detector.
\end{itemize}
This algorithm allows us to cover up for some of the limitations that the Detectron has, and we show that we achieve slight improvements over the Detectron and Part Affinity Field baselines with absolutely zero retraining. We keep no extra overlap because the person detector messes up in cases where the people are intertwined, and the box has to expand to cover up two people. The approach is simple, and requires zero retraining from baseline models. 
\subsection{Results}
We run this algorithm, keeping pre-trained Detectron and Part Affinity fields baselines. Note that the skeleton structure of Detectron and PAFs are different (COCO v/s MPII). We ran on COCO keypoints 2017 validation dataset, and MPII validation dataset. The results are positive but not very significantly high. This brings us to its limitations.

\subsection{Limitations}
The single big limitation is that it still doesn’t leverage the power of the Part Affinity fields to the fullest, in the sense that the detections of PAFs are not used, and the direction fields are used only to expand the bounding box. We are simply using the Detectron and its capabilities to find out poses, which may not be desirable when we are trying to find people in very constrained spaces.

\begin{figure}[t]
    \centering
    \includegraphics[width=\linewidth]{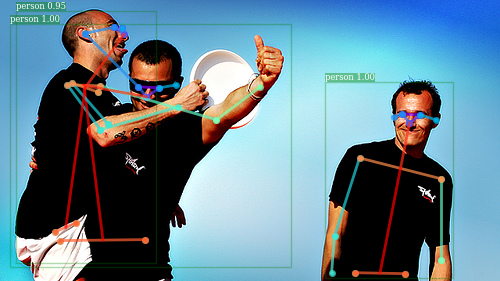}
    \caption{A very tough example}
    \label{fig:tough}
\end{figure}

\subsection{Future Work}
To tackle this, we are going to use a more flexible, and better method which doesn’t just depend on the bounding boxes to estimate a pose, and can use the PAF detections to make up for the missing detections of the RCNN. 

\section{Conclusion}
We saw some of the methods and approaches for human pose estimation. With the advent of powerful GPUs and scalability of deep networks, many methods focus on some sort of keypoint detection, and the association differs from method to method. CNNs provide us with complex, abstract features which are not possible by hand-engineering. However, it may be interesting to model human pose with more mathematical foundation, and replace the hand-engineered features with the newer, deep features. It may also be interesting to combine some of the methods (Region Proposal Networks and Part Affinity Fields, for example) to cover up each other’s mistakes and provide the best of both worlds.

\nocite{*}
{\small
\bibliographystyle{ieee}
\bibliography{egbib}
}

\end{document}